\documentclass[10pt,twocolumn,letterpaper]{article}

\usepackage[pagenumbers]{cvpr} 

\usepackage{placeins}

\makeatletter
\g@addto@macro\normalsize{%
  \setlength\abovedisplayskip{3pt}%
  \setlength\belowdisplayskip{3pt}%
  \setlength\abovedisplayshortskip{3pt}%
  \setlength\belowdisplayshortskip{3pt}%
}
\makeatother

\newcommand{\methodname}{DiaCap}
\newcommand{\eename}{DiaCap-E2E}
\newcommand{\benchname}{DiaDemBench-NoSub}
\newcommand{\predset}{Predicted-transcript}
\newcommand{\gtset}{GT-transcript}
\newcommand{\asrmodel}{\boldsymbol{\Psi}}
\newcommand{\capmodel}{\boldsymbol{\Theta}}
\newcommand{\mname}[1]{\texttt{#1}}

\usepackage{colortbl}
\usepackage{multirow}
\usepackage{pifont}
\newcommand{\cmark}{\ding{51}}
\newcommand{\xmark}{\ding{55}}

\definecolor{cvprblue}{rgb}{0.21,0.49,0.74}
\usepackage[pagebackref,breaklinks,colorlinks,allcolors=cvprblue]{hyperref}

\def\confName{CVPR}
\def\confYear{2027}

\title{DiaCap: Jointly Learning Who Speaks When for\\Audiovisual Dialogue Captioning}

\author{
    \addtocounter{footnote}{1}%
    Binfei Ji$^{1}$\thanks{Equal contribution.}\phantom{\textsuperscript{$\dagger$}}\thanks{Work done during an Adobe internship.}
    \hspace{0.7cm} Zhiwen Cao$^{2}$\footnotemark[2]
    \hspace{0.7cm} Anqi Yang$^{2}$\footnotemark[2]
    \hspace{0.7cm} Prem Seetharaman$^{2}$
    \hspace{0.7cm} Shuo Yang$^{2}$ \\
    \hspace{0.7cm} Ratheesh Kalarot$^{2}$
    \hspace{0.7cm} Yingyan (Celine) Lin$^{1}$
    \hspace{0.7cm} Mingze Xu$^{2}$ \\[.5ex]
    $^{1}$Georgia Tech \hspace{0.9cm} $^{2}$Adobe Research \\[.5ex]
    {\tt\small \{zhiwenc, anqiy, mingzex\}@adobe.com}
}

\begin{document}
\maketitle
\begin{abstract}
We present \methodname{}, an audiovisual LLM for comprehensive dialogue captioning, which produces the transcript, timestamps, and a visual description of the speaker for every utterance in a video. Considering that utterance timing and audiovisual diarization are tightly coupled, \methodname{} unifies the tasks in a single model, trained in two complementary stages. First, we train \methodname{} on 225K curated clips, using synthetic captions that densely label every utterance in detail. Second, we further refine the model via reinforcement learning on 2.9K human-corrected clips, using fine-grained feedback that directly evaluates utterance timing, speaker grouping, and speaker binding. On DiaDemBench, which we extend with utterance-level timestamps and subtitle redaction, both \methodname{} with off-the-shelf transcripts and its end-to-end variant \eename{} achieve strong results. Specifically, \methodname{} performs on par with Gemini~4~Argon in audiovisual diarization while clearly outperforming it in timestamp accuracy (76.59 \vs 42.76 TIME). It also maintains strong diarization on multilingual and minute-long videos, and transfers to audiovisual question answering.
\end{abstract}

\section{Introduction}
\label{sec:intro}

\begin{figure}[!t]
    \centering
    \includegraphics[width=\columnwidth]{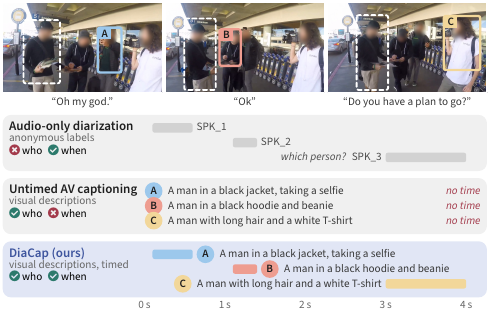}
    \vspace{-1.5em}
    \caption{\textbf{Comprehensive dialogue captioning.}
    \methodname{} describes \emph{who} says \emph{what}, and \emph{when}, in a single caption. Whereas audio-only diarization assigns each speaker only an anonymous ID, \methodname{} identifies each speaker with a detailed description.}
    \label{fig:teaser}
    \vspace{-1em}
\end{figure}

Comprehensive dialogue captioning requires more than accurate speech transcription: a model must also localize \emph{when} each utterance occurs and identify \emph{who} speaks it (\cref{fig:teaser}).
Unlike conventional speaker diarization~\cite{landini2022vbx,fujita2019eend,bredin2023pyannote}, which assigns anonymous labels even when it uses visual cues~\cite{chung2019whosaidthat,xu2022ava,liu2022msdwild}, \emph{audiovisual diarization} describes each speaker by detailed visual appearance.
However, general-purpose audiovisual LLMs~\cite{xu2025qwen3omni,qwen2026qwen35omni,google2026gemini31,google2026gemini4} and existing dialogue captioners~\cite{chen2026diadem,tang2026dorca} have yet to reliably determine \emph{when} and \emph{who} together, particularly in multi-person scenes or with speakers moving in and out of view.

\begin{figure*}[!t]
    \centering
    \includegraphics[width=\textwidth]{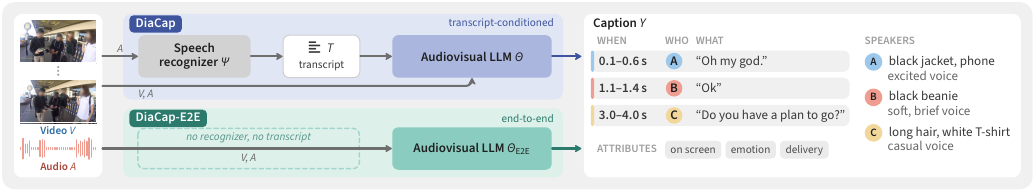}
    \vspace{-1.5em}
    \caption{\textbf{Overview of \methodname{}.}
    A speech recognizer $\asrmodel$ transcribes the audio, and an audiovisual LLM $\capmodel$ takes the transcript $T$ together with the video and audio to predict the timestamps and speaker of every utterance.
    \eename{} predicts $Y$ directly without a recognizer.}
    \label{fig:framework}
    \vspace{-1em}
\end{figure*}

We argue that this challenge stems from learning utterance timing and speaker attribution separately, even though they are tightly coupled.
For example, knowing who is speaking helps locate utterance boundaries, while accurate boundaries are needed to bind each utterance to the right speaker.
We present \textbf{\methodname{}}, an audiovisual LLM for comprehensive dialogue captioning (\cref{fig:framework}) that unifies both tasks in a single model by learning from two complementary signals.
First, \emph{dense supervision} provides complete captions that label every utterance with its transcript, timestamps, and speaker, and describe each speaker by visual appearance.
Second, as a more direct signal, \emph{fine-grained feedback} evaluates how accurately each utterance is timed and whether each speaker is correctly identified.

To provide dense supervision at scale, we first train \methodname{} with supervised fine-tuning (SFT) on 225K clips, curated from 1.25M raw clips to ensure audiovisual correspondence and a balanced data distribution.
Since manual annotation is impractical at this scale, we label these clips with a synthetic annotation pipeline that combines the strengths of two models.
An audiovisual LLM~\cite{google2026gemini31,qwen2026qwen35omni} yields accurate transcripts and diarization but only coarse timestamps, whereas a speech aligner~\cite{kumar2026tac} localizes speech precisely but cannot see the speakers.
We retain the speaker-attributed utterances from the audiovisual LLM and refine their timestamps with the aligner.
Because the pipeline chains several models and is thus complex and costly to run, we use it only as an offline data engine.

The next-token objective of SFT imitates the reference caption token by token, without directly measuring the accuracy of the predicted timestamps and speaker assignments.
We therefore further optimize \methodname{} via reinforcement learning (RL)~\cite{yu2025dapo}, using fine-grained feedback as the reward.
The RL data consists of about 2.9K human-corrected clips mined from visually challenging scenarios, such as multi-person scenes, small or blurry faces, and delayed speaker appearances.
Instead of a single score that conflates different error types, our reward decomposes into three terms.
A \emph{timestamp reward} scores each predicted utterance boundary by its distance from the reference.
For diarization, a description-agnostic \emph{grouping reward} checks whether each person's utterances share a single predicted speaker, penalizing both splitting one person and merging several people.
A \emph{binding reward} then asks a multimodal judge whether each predicted description refers to the right person, averaging over speakers so that minor speakers count as much as dominant ones.

Considering that transcription rarely relies on visual cues~\cite{xu2025qwen3omni,elevenlabs2026scribe,microsoft2026maitranscribe}, \methodname{} takes off-the-shelf transcripts by default and focuses on diarization and timestamping.
We also train \eename{}, an end-to-end variant that outputs the complete dialogue caption directly.
We evaluate on DiaDemBench~\cite{chen2026diadem}, which we extend with utterance-level timestamps and subtitle redaction for more rigorous assessment.
Experimental results show that both models perform strongly, with \methodname{} using transcripts from an off-the-shelf recognizer and \eename{} predicting its own.
Specifically, \methodname{} performs on par with Gemini~4~Argon~\cite{google2026gemini4} in audiovisual diarization while substantially outperforming it in timestamp accuracy (76.59 \vs 42.76 TIME).
Beyond DiaDemBench, \methodname{} maintains strong diarization on multilingual videos, including languages underrepresented in its training data, and on minute-long videos~\cite{tang2026dorca}. It also shows promising transfer to audiovisual question answering~\cite{nguyen2026see}.

\section{Related Work}
\label{sec:related}

\noindent\textbf{Speaker diarization} predicts \textit{who} speaks each word or utterance.
Audio-only methods partition speech by speaker using voice characteristics alone~\cite{anguera2012review,park2022review}, through embedding clustering~\cite{landini2022vbx,snyder2018xvector,wang2018lstm,zhang2019uisrnn}, end-to-end prediction~\cite{fujita2019eend,horiguchi2020eda,plaquet2023powerset}, hybrid systems~\cite{bredin2023pyannote,kinoshita2021integrating}, or joint recognition and diarization~\cite{elshafey2019joint,kanda2020sot,kanda2022t2d,mosi2026moss,park2025sortformer,peng2026vibevoice,yin2025speakerlm,wang2024diarizationlm}.
We focus on audiovisual diarization, which must also link each utterance to a visible person.
Audiovisual methods associate speech with faces via synchrony~\cite{chung2016syncnet,cutler2000look,iashin2024synchformer,owens2018multisensory} and active speaker detection~\cite{kopuklu2021asdnet,alcazar2020asc,liao2023light,min2022spell,roth2020ava,tao2021talknet,wang2024loconet} to diarize~\cite{brown2021facebodyvoice,cheng2025avt,chung2019whosaidthat,gebru2018avdiar,huang2026cinesrd,liu2022msdwild,xu2022ava}, but still output anonymous clusters or face tracks.
Methods that name speakers require prior identity information, either cast galleries~\cite{everingham2006buffy,nagrani2017sherlock,korbar2024look,huh2024context} or predefined appearance descriptions~\cite{bai2026humanomnispeaker}.
\methodname{} instead generates a visual description of each speaker from the video and links every utterance to that person.

\noindent\textbf{Temporal localization} determines \textit{when} each word or utterance is spoken.
Timestamps are typically obtained by aligning transcripts to audio~\cite{bain2023whisperx,kurzinger2020ctc,mcauliffe2017mfa}, including with LLM-based forced aligners~\cite{shi2026qwen3asr}, or generated by audio-language models as timestamped captions~\cite{kumar2026tac} or word-level annotations of given transcripts~\cite{kons2026timestamps}.
In video, moment retrieval~\cite{gao2017tall,hendricks2017mcn,lei2021qvhighlights} and dense event captioning~\cite{krishna2017dense,yang2023vid2seq} have led to video and audiovisual LLMs that predict temporal boundaries~\cite{chowdhury2024meerkat,huang2024vtimellm,ren2024timechat,tang2025avicuna,wang2025timer1}, but these localize queried moments or scene events rather than individual utterances.
Only a few audiovisual LLMs output utterance-level timestamps~\cite{tang2026dorca,google2026gemini4,qwen2026qwen35omni}, and these remain coarse.
\methodname{} targets this utterance-level setting, predicting the boundaries of each utterance together with its speaker.

\noindent\textbf{Audiovisual dialogue captioning.}
General-purpose audiovisual LLMs, including Qwen-Omni~\cite{qwen2026qwen35omni,xu2025qwen25omni,xu2025qwen3omni}, Gemini~\cite{comanici2025gemini25,google2026gemini4,google2026gemini31}, and others~\cite{cheng2024videollama2,fu2024vita,sun2024videosalmonn,mimo2026v26flash,ye2025omnivinci,zhang2023videollama}, can generate transcripts, speaker descriptions, and timestamps from synchronized video and audio.
Detailed audiovisual captioners improve content coverage and audiovisual correspondence~\cite{chen2023vast,chen2025avocado,ge2025arc,liu2025valor,ma2025omnicaptioner,tang2025videosalmonn,wu2025ugc,shi2026evacap,wu2026avcap}, and structured script generators organize scenes, characters, events, and dialogue into timestamped outputs~\cite{pu2026omniscript,hunyuan2026script,yao2026timechatcaptioner}.
Most relevant to us are the dialogue captioners, such as DiaDem~\cite{chen2026diadem} and D-ORCA~\cite{tang2026dorca}, whose speaker attribution still trails strong general-purpose audiovisual LLMs. Moreover, DiaDem does not output timestamps, and D-ORCA outputs only whole-second ones.
\methodname{} fills this gap, which matches the speaker attribution of the strongest general-purpose audiovisual LLMs while producing the most accurate utterance timestamps.

\section{Dialogue Captioning}
\label{sec:method}

Given a video $V$ and its audio track $A$, dialogue captioning aims to produce a caption $Y$ consisting of a list of utterances and a list of speakers (\cref{fig:framework}).
Each utterance is a continuous segment of speech from a single speaker, annotated with its transcript, start and end timestamps, and speaker label.
The label points to an entry in the speaker list, where each speaker is described by visual appearance (\eg, ``a man in a blue shirt'') so that they can be matched to a person in the video.
In this paper, we refer to predicting the utterance-level timestamps as utterance timing, and to building the speaker list and assigning each utterance to a speaker as audiovisual diarization.

\subsection{Pipeline Overview}
\label{sec:pipeline}

\noindent\textbf{Transcript-conditioned pipeline.}
Transcription is well served by strong off-the-shelf speech recognizers~\cite{xu2025qwen3omni,elevenlabs2026scribe,microsoft2026maitranscribe,google2026gemini4}, whereas timestamps and diarization remain the main bottlenecks of current dialogue captioning.
We therefore separate transcription from the rest of the task, which lets us reuse these recognizers and focus our model on predicting audiovisual diarization and precise timestamps.

A speech recognizer $\asrmodel$ produces the transcript, and an audiovisual LLM $\capmodel$ predicts the timestamps and diarization:
\begin{equation}
  T=\asrmodel(A)
  \quad\text{and}\quad
  Y=\capmodel(V,A,T).
  \label{eq:diacap}
\end{equation}
We refer to this pipeline as \textbf{\methodname{}} (upper panel in \cref{fig:framework}).
The transcript $T$ is a single text that covers all speech in the clip, without utterance boundaries, timestamps, or speaker labels.
Given $T$, the model $\capmodel$ splits it into utterances, predicts their timestamps, and diarizes them.
Since $\capmodel$ only needs text from $\asrmodel$, any speech recognizer or existing human transcript can provide $T$, and \methodname{} benefits from better transcripts without retraining $\capmodel$ (\cref{sec:ablations}).

\noindent\textbf{End-to-end pipeline.}
We also train a single model $\capmodel_{\text{E2E}}$ that predicts the transcript, timestamps, and diarization directly from the video and audio:
\begin{equation}
  Y=\capmodel_{\text{E2E}}(V,A).
  \label{eq:e2e}
\end{equation}
We refer to it as \textbf{\eename{}} (lower panel in \cref{fig:framework}).
It shares the same backbone, training data, and output format as \methodname{}, but does not take $T$ as input.
This shows that our framework also applies without an external recognizer.

\subsection{Supervised Fine-Tuning}
\label{sec:sft}

\begin{figure*}[!t]
    \centering
    \includegraphics[width=\textwidth]{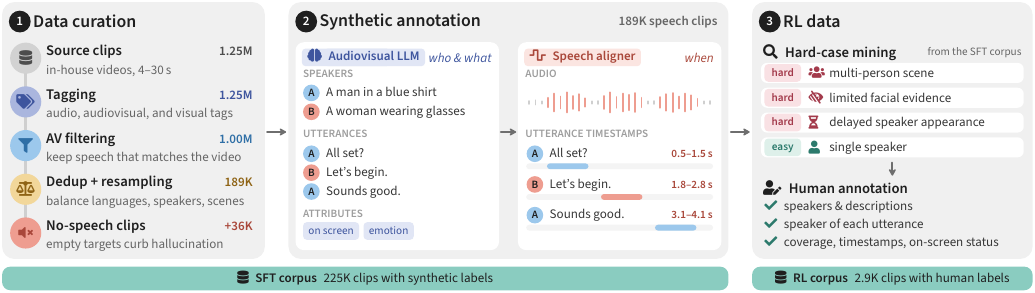}
    \vspace{-1.5em}
    \caption{\textbf{Training data curation and annotation.}
    (1)~Audio, audiovisual, and visual experts tag each clip.
    We remove clips with unrelated or out-of-sync audio, deduplicate and resample the rest to balance language, speaker count, and visual scenes, and add no-speech clips with empty targets.
    (2)~An audiovisual LLM annotates who says what, and a speech aligner timestamps each utterance.
    (3)~Human annotators correct hard clips mined from the SFT corpus, plus fewer single-speaker clips, to form the RL corpus.}
    \label{fig:data_pipeline}
    \vspace{-1em}
\end{figure*}

SFT requires a large and diverse set of videos with complete dialogue captions (\ie, transcripts, timestamps, and audiovisual diarization).
To the best of our knowledge, no public dataset provides such annotations at scale, and no single existing model produces them reliably.
We therefore build the SFT corpus from raw videos through data curation and synthetic annotation (\cref{fig:data_pipeline}).

\noindent\textbf{Data curation.}
We curate an audiovisually aligned and balanced set of clips in four steps (\cref{fig:data_pipeline}, step 1).

\textit{Tagging.}
We start from an in-house collection of about 1.25M clips of 4 to 30s, most of which are cut from longer videos.
For the audio, \mname{pyannote.audio}~\cite{plaquet2023powerset,bredin2023pyannote} diarizes the speech and provides the number of speakers and the amount of overlapped speech, and the language is taken from the metadata of the collection.
For the audiovisual correspondence, Synchformer~\cite{iashin2024synchformer} predicts a relatedness score $r$, which indicates whether the sound is produced by the visible content, and the offset $\delta$ between the audio and the video.
For the visual content, we track and cluster faces to count the visible people, and LR-ASD~\cite{liao2025lrasd} detects which of them are speaking.

\textit{Audiovisual filtering.}
We keep a clip only if $r>0.3$ and $|\delta|<0.5$\,s, which removes clips whose audio is unrelated to the frames, such as voice-overs, and clips whose audio is out of sync so that lip motion cannot be matched to the speech.
The filter keeps about 80\% of the pool.
We do not require a visible or speaking face, since off-screen speakers are part of the task and are described by voice alone.

\textit{Deduplication and resampling.}
The filtered pool is highly imbalanced (\eg, English and Mandarin account for about 70\% of the clips, and only 16\% have three or more speakers).
We first remove duplicate windows from the same source video and cap their number of clips, so that a few long videos do not dominate.
We then draw a stratified sample that balances, in order of priority, the language over 18 target languages, the number of speakers, the clip duration, the amount of overlapped speech, and the numbers of visible and speaking faces.
The last two factors retain the rare clips with overlapping speech or several visible people, on which diarization is most difficult.

\textit{No-speech clips.}
The zero-speaker stratum provides clips without speech, such as clips with only music, ambient sound, or silence.
These clips skip annotation and receive empty targets, so that the model learns not to hallucinate dialogue.
The final corpus contains about 225K clips (189K with speech and 36K without speech) with a total duration of 1,175 hours. Appendix~\ref{app:data} shows its distributions of speaker count and language.

\noindent\textbf{Synthetic annotation.}
Human annotation at this scale is costly and time-consuming, especially for diarization.
We therefore build a synthetic annotation pipeline that combines two complementary models (\cref{fig:data_pipeline}, step 2).
An audiovisual LLM (\eg, Gemini~3.1~Pro~\cite{google2026gemini31}) gives accurate transcripts and diarization, but its timestamps remain coarse even with careful prompting.
A speech aligner, TAC~\cite{kumar2026tac}, localizes speech precisely but cannot see the speakers.
The audiovisual LLM first annotates the transcript, utterance segmentation, speaker list, speaker assignments, and attributes.
TAC then refines the timestamps of each utterance while keeping the rest of the annotation, and clips that fail the alignment are discarded.
This combination substantially improves TIME from 23.54 to 72.42 over Gemini~3.1~Pro alone, while keeping its transcripts and diarization (\cref{tab:main_results}).
We use the pipeline as an offline data engine, but chaining several models and a proprietary LLM makes it too expensive for everyday use.

\noindent\textbf{Learning objective.}
We build $\capmodel$ upon a pretrained audiovisual language model (\eg, Qwen3-Omni~\cite{xu2025qwen3omni}) and train it with standard next-token prediction.
Given the caption $Y$ as a token sequence $y_1,\dots,y_L$, the loss is
\begin{equation}
  \mathcal{L}_{\text{SFT}}=-\sum_{j=1}^{L}\log P_{\capmodel}\bigl(y_j\mid y_{<j},V,A,T\bigr),
  \label{eq:sft}
\end{equation}
and \eename{} is trained with the same loss without $T$.
If the base model writes a thinking trace before its answer, we prepend an empty thinking block (\texttt{<think></think>}) to every target so that the model outputs the caption directly.

\subsection{{Reinforcement Learning}}
\label{sec:rl}

We further refine $\capmodel$ and $\capmodel_{\text{E2E}}$ with RL by providing direct training signals for timestamps and audiovisual diarization.

\noindent\textbf{Hard-case mining and annotation.}
We focus on three types of visually difficult scenes, namely
(1)~\emph{multi-person scenes}, where several faces remain visible while only one person speaks,
(2)~\emph{limited visual evidence}, where the speaker's mouth movements are difficult to observe due to low resolution, blur, occlusion, or unfavorable viewing angles; and
(3)~\emph{delayed speaker appearance}, where speech begins while a non-speaking person is shown on screen, and the actual speaker only becomes visible later.
We retrieve clips of each type from the SFT corpus with rules on their video tags and synthetic annotations (\eg, face and speaker counts), and add a smaller set of single-speaker clips to avoid biasing the policy toward extra speakers, which yields about 2.9K clips (\cref{fig:data_pipeline}, step 3).
To obtain reliable references $Y^\ast$ on these clips, human annotators correct the synthetic annotation of each clip against the video, fixing the speaker list, reassigning utterances, completing missing speech, and adjusting timestamps.

\noindent\textbf{Reward design.}
A sampled caption $\hat{Y}$ and the reference $Y^\ast$ may split the speech into different utterances, so we first match them at the word level.
Since both captions contain nearly the same words, we align their word sequences by edit distance, using characters instead of words for CJK languages.
Each reference word is thus mapped to the predicted utterance that contains it, which gives its predicted speaker and timestamps.

\textit{Timestamp reward.}
For each start and end time of a reference utterance, we take the corresponding predicted boundary and assign a credit of one if the error is within 0.3s, decreasing linearly to zero at 1s.
A prediction may merge two consecutive reference utterances and thus miss the boundary between them.
We give zero credit to such a missing boundary only if it marks a speaker change or a pause of at least 0.5s, so that merging continuous speech of one speaker is not penalized.
Extra predicted boundaries inside a reference utterance are not scored.
$R_{\text{time}}$ is the mean credit over all scored boundaries.

\textit{Diarization rewards.}
We evaluate diarization with two rewards:
the grouping reward checks whether the utterances of each person are grouped under one predicted speaker, regardless of the descriptions, and the binding reward checks whether each predicted speaker is described as the right person.
Let $\mathcal{G}$ and $\mathcal{P}$ be the reference and predicted speakers, and let $n_g$ be the number of utterances of reference speaker $g$, with $N=\sum_{g}n_g$.
We count in $C_{pg}$ the utterances of $g$ that are assigned to predicted speaker $p$.
An utterance split across several predicted speakers is counted fractionally by its aligned words, and utterances spoken by several people at once are excluded.
For grouping, we match each reference speaker to a distinct predicted speaker with the one-to-one mapping $\pi$ that covers the most utterances:
\begin{equation}
  R_{\text{grp}}=\frac{1}{N}\max_{\pi}\sum_{g\in\mathcal{G}}C_{\pi(g)\,g}.
  \label{eq:grouping}
\end{equation}
Splitting one person into several speakers and merging several people into one both leave utterances unmatched and thus lower the reward.
For binding, a multimodal judge watches the clip and sets $J_{pg}=1$ if descriptions $p$ and $g$ refer to the same person and $J_{pg}=0$ otherwise:
\begin{equation}
  R_{\text{bind}}=\frac{1}{|\mathcal{G}|}\sum_{g\in\mathcal{G}}\frac{1}{n_g}\sum_{p\in\mathcal{P}}C_{pg}\,J_{pg}.
  \label{eq:binding}
\end{equation}
It averages over reference speakers, so a speaker with few utterances counts as much as the main speaker, unlike the utterance-weighted REF metric of DiaDem~\cite{chen2026diadem}.

The final reward is
\begin{equation}
  R=\lambda_{\text{time}}R_{\text{time}}+\lambda_{\text{grp}}R_{\text{grp}}+\lambda_{\text{bind}}R_{\text{bind}}+\lambda_{\text{asr}}R_{\text{asr}},
  \label{eq:reward}
\end{equation}
where $R_{\text{asr}}=1-\min(\mathrm{WER},1)$ scores the predicted transcript against the reference.
DiaCap receives the transcript as input, so we set $\lambda_{\text{asr}}=0$, $\lambda_{\text{time}}=0.2$, and $\lambda_{\text{grp}}=\lambda_{\text{bind}}=0.4$.
DiaCap-E2E writes its own transcript, so we add $\lambda_{\text{asr}}=0.2$ and lower $\lambda_{\text{grp}}$ to $0.2$, keeping $\lambda_{\text{time}}=0.2$ and $\lambda_{\text{bind}}=0.4$.
Captions that cannot be parsed receive zero reward.

\noindent\textbf{Policy optimization.}
We optimize $\capmodel$ and $\capmodel_{\text{E2E}}$ with DAPO~\cite{yu2025dapo}, initialized from the SFT model.
For each clip, we sample a group of eight captions and compute the advantage of each caption as its reward minus the group mean, which raises the likelihood of captions that outperform the group average.
Following Dr.~GRPO~\cite{liu2025understanding}, we do not normalize the advantages by the group standard deviation, since this normalization overweights clips whose captions differ only slightly in reward.
As in DAPO, we discard groups with identical rewards, which provide no learning signal, and apply a soft penalty to overlong responses.

\section{Experiments}
\label{sec:experiments}

\begin{table*}[!t]
    \centering
    \footnotesize
    \setlength{\tabcolsep}{3.5pt}
    \begin{tabular}{@{}l *{2}{w{c}{2.6em}>{\color{black!55}}w{c}{2.6em}} w{c}{2.6em}>{\color{black!55}}w{c}{2.6em} c *{2}{w{c}{2.6em}>{\color{black!55}}w{c}{2.6em}}@{}}
        \toprule
        & \multicolumn{6}{c}{\predset{}} & & \multicolumn{4}{c@{}}{\gtset{}} \\
        \cmidrule(lr){2-7}\cmidrule(l){9-12}
        & \multicolumn{2}{c}{REF$\uparrow$} & \multicolumn{2}{c}{ASR$\uparrow$} & \multicolumn{2}{c}{TIME$\uparrow$} & & \multicolumn{2}{c}{REF$\uparrow$} & \multicolumn{2}{c@{}}{TIME$\uparrow$} \\
        \cmidrule(lr){2-3}\cmidrule(lr){4-5}\cmidrule(lr){6-7}\cmidrule(lr){9-10}\cmidrule(l){11-12}
        Method / Subtitles & \xmark & \cmark & \xmark & \cmark & \xmark & \cmark & & \xmark & \cmark & \xmark & \cmark \\
        \midrule
        \multicolumn{12}{@{}l}{\textit{Proprietary}} \\
        Annotation pipeline & 69.18 & 71.31 & 76.57 & 80.56 & \underline{72.42} & \underline{74.49} & & - & - & - & - \\
        Qwen3.5-Omni-Plus & 66.78 & 66.51 & 77.04 & 78.15 & 22.17 & 24.87 & & 71.32 & 71.87 & 28.29 & 31.26 \\
        Gemini 3.1 Pro & 69.18 & 71.31 & 76.57 & 80.56 & 23.54 & 23.81 & & 80.08 & 79.82 & 32.27 & 32.11 \\
        Gemini 4 Argon~\cite{google2026gemini4} & \underline{70.93} & \underline{73.56} & \underline{77.41} & \underline{81.85} & 42.76 & 44.80 & & \underline{80.87} & \underline{80.67} & \underline{48.08} & \underline{48.83} \\
        \midrule
        \multicolumn{12}{@{}l}{\textit{Open-weight}} \\
        ARC-Qwen-Video-Narrator~\cite{ge2025arc} & 33.90 & 31.91 & 49.75 & 46.48 & - & - & & - & - & - & - \\
        HumanOmniV2~\cite{yang2025humanomniv2} & 15.37 & 14.97 & 23.03 & 22.14 & - & - & & - & - & - & - \\
        MiMo-V2.6-Flash~\cite{mimo2026v26flash} & 58.04 & - & 68.67 & - & 17.96 & - & & 72.62 & - & 18.42 & - \\
        Qwen2.5-Omni~\cite{xu2025qwen25omni} & 31.50 & 30.92 & 55.10 & 54.11 & \phantom{0}4.65 & \phantom{0}4.71 & & 40.05 & 40.11 & \phantom{0}9.31 & \phantom{0}8.82 \\
        UGC-VideoCaptioner~\cite{wu2025ugc} & 24.93 & 24.57 & 37.11 & 37.54 & - & - & & 34.11 & 34.49 & \phantom{0}5.23 & \phantom{0}4.99 \\
        AVoCaDO~\cite{chen2025avocado} & 42.44 & 40.11 & 53.80 & 51.85 & - & - & & - & - & - & - \\
        D-ORCA~\cite{tang2026dorca} & 50.97 & 51.29 & 63.09 & 66.91 & 11.24 & 11.84 & & - & - & - & - \\
        Qwen3-Omni-Thinking~\cite{xu2025qwen3omni} & 52.76 & 49.88 & 74.29 & 71.50 & 22.52 & 18.40 & & 55.28 & 55.14 & 18.81 & 17.41 \\
        DiaDem~\cite{chen2026diadem} & 63.79 & 65.87 & 75.62 & \textbf{79.46} & - & - & & - & - & - & - \\
        \midrule
        \rowcolor{gray!12}
        \methodname{} (SFT) & 68.58 & 69.44 & 78.38 & 78.50 & 75.69 & 75.55 & & 79.11 & 79.15 & 86.42 & 86.06 \\
        \rowcolor{gray!12}
        \methodname{} (SFT + RL) & \textbf{71.70} & \textbf{71.87} & {78.99} & 78.97 & {76.59} & \textbf{76.56} & & \textbf{81.05} & \textbf{81.01} & \textbf{88.76} & \textbf{89.31} \\
        \rowcolor{gray!12}
        \methodname{}-E2E (SFT) & 70.60 & 71.72 & 78.98 & 81.76 & 74.86 & 75.44 & & - & - & - & - \\
        \rowcolor{gray!12}
        \methodname{}-E2E (SFT + RL) & 71.59 & \textbf{73.02} & \textbf{79.55} & \textbf{82.13} & \textbf{76.64} & \textbf{77.83} & & - & - & - & - \\
        \bottomrule
    \end{tabular}
    \vspace{-0.5em}
    \caption{\textbf{Comparison with state-of-the-art models.} Black columns (\xmark) are evaluated on \benchname{}, in which burned-in subtitles are redacted, and \textcolor{black!55}{gray columns (\cmark)} on the original videos with subtitles, using the same annotations. \textbf{Bold} and \underline{underline} mark the best open-weight and proprietary results, respectively.}
    \label{tab:main_results}
    \vspace{-1em}
\end{table*}

\subsection{Settings}
\label{sec:setup}

\noindent\textbf{Benchmarks.}
We evaluate on DiaDemBench~\cite{chen2026diadem}, which contains 1,039 video clips whose utterances are each annotated with a transcript and a natural-language description of the speaker.
Since DiaDem provides no timestamps, we manually annotate the start and end times of every utterance, and \emph{DiaDemBench} hereafter refers to the benchmark with these timestamps.
To prevent models from exploiting the burned-in subtitles for transcription, we further detect the subtitles in every frame with PP-OCRv5~\cite{cui2026pp} and blur the detected regions, leaving the audio and annotations unchanged, which affects 598 of the 1039 clips.
We refer to this subtitle-redacted version as \emph{\benchname{}} and report all results on it unless noted otherwise.

\noindent\textbf{Evaluation metrics.}
\label{sec:metrics}
We adopt the REF and ASR metrics of DiaDemBench~\cite{chen2026diadem} and introduce TIME to evaluate timestamps.
\emph{REF} measures whether each utterance is attributed to the correct speaker, as judged by Gemini-2.5-Flash.
\emph{ASR} measures the transcription accuracy of the matched utterances.
\emph{TIME} measures whether both the start and end times of each matched utterance are within 0.3s of the reference.

\noindent\textbf{Baselines.}
We compare \methodname{} with state-of-the-art proprietary and open-weight models, all evaluated by us with their official implementations.
General-purpose audiovisual LLMs (\eg, Gemini~3.1~Pro) receive our prompt, which requests every field of the caption.
Dialogue captioners (\eg, DiaDem) use their official prompts, and a text-only extractor~\cite{comanici2025gemini25} converts their free-form captions into dialogues.
We also report the annotation pipeline that produces the synthetic labels of \methodname{} (\cref{sec:sft}).

\noindent\textbf{Evaluation protocol.}
We report results under two settings.
In the \emph{\predset{}} setting, no ground-truth text is given, so the baselines transcribe the speech themselves, whereas \methodname{} takes the transcript from an off-the-shelf speech recognizer.
In the \emph{\gtset{}} setting, models additionally receive the ground-truth transcript as a single unsegmented text, which removes transcription errors and isolates diarization and timestamps.

\subsection{Implementation Details}
\label{sec:implementation}

\noindent\textbf{Model architecture.}
We build \methodname{} and \eename{} on Qwen3-Omni-30B-A3B-Thinking~\cite{xu2025qwen3omni}.
We apply LoRA~\cite{hu2021lora} with rank 512 and $\alpha=1024$ to the attention projections of the LLM, and freeze the experts, the router, and the audio and vision encoders.
Each video is resized so that its short edge is at most 360 pixels, preserving its aspect ratio, and is sampled at 4 frames per second.

\noindent\textbf{SFT training.}
We train $\capmodel$ and $\capmodel_{\text{E2E}}$ on the 225K clips labeled by our synthetic pipeline, which uses Gemini~3.1~Pro~\cite{google2026gemini31} as its audiovisual LLM. The data is formatted correspondingly to the required input output of both models.
We use AdamW with $\beta_1=0.9$ and $\beta_2=0.95$, and a peak learning rate of $6\times10^{-5}$ with 100 warmup steps and cosine decay over 2 epochs.
Training uses a global batch size of 128 on 16 NVIDIA B200 GPUs, and we select the checkpoint after 1.5 epochs.

\noindent\textbf{RL training.}
After SFT, we further train $\capmodel$ and $\capmodel_{\text{E2E}}$ with DAPO~\cite{yu2025dapo} on the 2.9K human-corrected clips.
We use a global batch size of 32 clips and each clip has 8 rollouts. We train with a learning rate of $10^{-5}$ for 100 steps for $\capmodel$ and 60 steps for $\capmodel_{\text{E2E}}$.
The binding reward uses Gemini-2.5-Flash~\cite{comanici2025gemini25} as its judge.

\noindent\textbf{Inference.}
In the \predset{} setting, \methodname{} uses Qwen3-Omni-Captioner~\cite{xu2025qwen3omni} as its speech recognizer $\asrmodel$ unless noted otherwise, whereas \eename{} predicts the transcript itself.
In the \gtset{} setting, \methodname{} directly takes the GT transcript instead.

\begin{figure*}[!t]
    \centering
    \includegraphics[width=\textwidth]{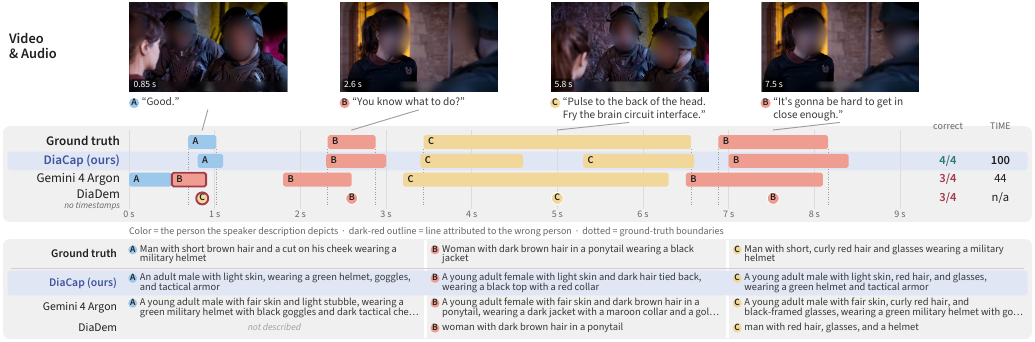}
    \vspace{-1.5em}
    \caption{\textbf{Qualitative comparison with similar-looking speakers.}
    Each track shows one model's predicted utterances over time, colored by the ground-truth speaker that its description depicts, with wrong speakers outlined in dark red.}
    \label{fig:case_study}
    \vspace{-1em}
\end{figure*}

\subsection{Main Results}
\label{sec:main_results}

\cref{tab:main_results} compares \methodname{} with state-of-the-art models. We discuss the results on \benchname{} (\xmark{} columns).
\noindent\textbf{Comparison with open-weight models.}
DiaDem, the best open-weight model, does not predict timestamps, and \methodname{} outperforms it by $+$7.91 REF, with \methodname{}-E2E close behind.
Among the models that predict timestamps, the best TIME is only 22.52 (Qwen3-Omni-Thinking), while both of our models exceed 76.

\noindent\textbf{Comparison with proprietary models.}
In the \predset{} setting, both of our models outperform Gemini~4~Argon in all three metrics, most notably by over $+$33 TIME.
\methodname{} also outperforms the annotation pipeline that produces its SFT labels by $+$2.52 REF, $+$2.42 ASR, and $+$4.17 TIME, and \methodname{}-E2E by similar margins.

\noindent\textbf{\gtset{} setting.}
We evaluate only \methodname{}, as \methodname{}-E2E does not take a transcript as input.
\methodname{} matches Gemini~4~Argon in REF (81.05 \vs 80.87), leads by $+$40.68 TIME, and outperforms its base model by $+$25.77 REF.
With GT transcripts, every model's REF rises, so transcription quality strongly affects diarization.

\noindent\textbf{Effect of RL.}
RL lifts the \predset{} REF of \methodname{} by $+$3.12, from below Gemini~3.1~Pro to the highest among all models.
\methodname{}-E2E starts from a stronger SFT model and gains less in REF, but RL improves all three of its metrics and gives the best open-weight ASR and TIME.

\noindent\textbf{Qualitative results.}
\cref{fig:case_study} shows an example with three speakers, two of whom wear similar military helmets.
Gemini~4~Argon misattributes the utterance and hallucinates an extra one, and DiaDem describes only two of the three speakers.
\methodname{} distinguishes the two similar-looking speakers, attributing and timing all four utterances correctly (clip-level TIME of 100 \vs 44 for Gemini~4~Argon).
More qualitative examples are provided in Appendix~\ref{app:qualitative}.

\subsection{Ablation Study}
\label{sec:ablations}
\begin{table}[!tbp]
    \centering
    \footnotesize
    \setlength{\tabcolsep}{2.5pt}
    \begin{tabular}{llrrr}
        \toprule
        Method & Transcript & REF$\uparrow$ & ASR$\uparrow$ & TIME$\uparrow$ \\
        \midrule
        \eename{} & Self-predicted & 71.59 & 79.55 & 76.64 \\
        \midrule
        & Gemini 4 Argon & 70.76 & 78.23 & 74.68 \\
        & Scribe v2 & 71.11 & 78.37 & 76.25 \\
        & \cellcolor{gray!15}Qwen3-Omni-Captioner$^*$ & \cellcolor{gray!15}71.70 & \cellcolor{gray!15}78.99 & \cellcolor{gray!15}76.59 \\
        \multirow{-4}{*}{\methodname{}} & MAI-Transcribe-2 & \textbf{72.48} & \textbf{81.23} & \textbf{78.93} \\
        \bottomrule
    \end{tabular}
    \vspace{-0.5em}
    \caption{\textbf{Transcript-conditioned \vs end-to-end prediction.} \eename{} predicts the transcript itself; \methodname{} takes transcripts from speech recognizers without retraining. Both are trained with SFT + RL. $^*$Default recognizer.}
    \label{tab:joint}
    \vspace{-1em}
\end{table}

Unless noted otherwise, all ablations use the \predset{} setting on \benchname{}.

\noindent\textbf{Effect of subtitles.}
We also evaluate all models on the original DiaDemBench videos with burned-in subtitles (\cmark{} columns in \cref{tab:main_results}).
Many baselines improve markedly in the \predset{} setting by reading the dialogue from the subtitles (\eg, $+$4.44 ASR and $+$2.63 REF for Gemini~4~Argon), and \methodname{}-E2E, which also transcribes speech itself.
These gains disappear with GT transcripts (\eg, 80.87 \vs 80.67 REF for Gemini~4~Argon), confirming that the subtitles mainly help transcription.
In contrast, \methodname{} changes by less than 0.2 in all metrics, as it takes the audio-only off-the-shelf transcript.
We therefore recommend \benchname{}, a stricter but fairer setting that requires transcribing speech rather than exploiting subtitles.

\noindent\textbf{Transcript conditioning.}
We compare \methodname{} with \eename{}, both trained with SFT + RL, to test how an off-the-shelf transcript affects the caption (\cref{tab:joint}).
With the default Qwen3-Omni-Captioner, the two models perform comparably: \eename{} achieves higher ASR ($+$0.56), while REF and TIME differ by less than 0.11.
Since \methodname{} takes the transcript as input, it can use other speech recognizers~\cite{xu2025qwen3omni,elevenlabs2026scribe,microsoft2026maitranscribe,google2026gemini4} without retraining.
Its TIME increases with the recognizer's ASR, from 74.68 to 78.93 as ASR rises from 78.23 to 81.23, and with MAI-Transcribe-2, the strongest recognizer, it reaches the best results in \cref{tab:joint}.

\noindent\textbf{RL data.}
We test whether hard-case mining and human review improve the RL data of \methodname{}, compared with randomly sampled clips and automatic labels (\cref{tab:rl_data}).
With the same automatic labels, mined clips outperform random ones by $+$0.43 REF and $+$0.79 TIME.
Human review of the mined labels adds a further $+$1.38 REF and $+$0.44 TIME, for a total gain of $+$1.81 REF and $+$1.23 TIME over random clips with automatic labels.

\begin{table}[!tbp]
    \centering
    \footnotesize
    \setlength{\tabcolsep}{3pt}
    \begin{tabular}{llrrr}
        \toprule
        RL data & Labels & REF$\uparrow$ & ASR$\uparrow$ & TIME$\uparrow$ \\
        \midrule
        Random & Automatic & 69.89 & 78.53 & 75.36 \\
        Mined & Automatic & 70.32 & 78.90 & 76.15 \\
        \rowcolor{gray!15}
        Mined & Human-reviewed & \textbf{71.70} & \textbf{78.99} & \textbf{76.59} \\
        \bottomrule
    \end{tabular}
    \vspace{-0.5em}
    \caption{\textbf{Effect of RL data on DiaCap.} All variants start from the same SFT checkpoint and use our reward as in \cref{eq:reward}.}
    \label{tab:rl_data}
    \vspace{-1em}
\end{table}

\begin{table}[!tbp]
    \centering
    \footnotesize
    \setlength{\tabcolsep}{3pt}
    \begin{tabular}{lrrr}
        \toprule
        Method & REF$\uparrow$ & ASR$\uparrow$ & TIME$\uparrow$ \\
        \midrule
        SFT & 68.58 & 78.38 & 75.69 \\
        \;+RL w/ $0.8\,\mathrm{REF}{+}0.2R_{\text{time}}$ & 70.40 & 78.86 & 75.92 \\
        \rowcolor{gray!15}
        \;+RL w/ $0.4R_{\text{grp}}{+}0.4R_{\text{bind}}{+}0.2R_{\text{time}}$ & \textbf{71.70} & \textbf{78.99} & \textbf{76.59} \\
        \bottomrule
    \end{tabular}
    \vspace{-0.5em}
    \caption{\textbf{Effect of the RL reward on DiaCap.} Both rewards are trained on the same human-reviewed mined clips and evaluated at RL step 100; the shaded row is our reward (\cref{eq:reward}).}
    \label{tab:reward}
    \vspace{-1em}
\end{table}

\noindent\textbf{RL reward design.}
We compare our decomposed reward with a baseline that replaces the grouping and binding rewards with REF~\cite{chen2026diadem} (\cref{tab:reward}) on \methodname{}.
Although the baseline directly optimizes REF, our reward outperforms it by $+$1.30 REF and $+$0.67 TIME, and improves REF over SFT by $+$3.12 instead of $+$1.82.
We attribute this mainly to the binding reward, which weights every speaker equally, whereas REF is dominated by frequent speakers and gives little credit for correctly attributing minor ones.

\subsection{Generalization}
\label{sec:generalization}

\noindent\textbf{Multilingual videos.}
\label{sec:multilingual}
Since DiaDemBench mostly covers English, Chinese, Japanese, and Korean, we curate a multilingual benchmark to evaluate \methodname{} on more languages.
It contains 140 clips in 14 languages, with burned-in subtitles redacted as in \benchname{}.
Ten of these languages are well represented in our SFT and RL data, whereas Vietnamese, Thai, Persian, and Burmese are underrepresented (at most 0.25\% of SFT data).
For the ground truth, we keep the clips on which Gemini, MAI-Transcribe-2, and Scribe~v2 agree on the transcript, and manually correct the speaker descriptions, assignments, and timestamps.
We evaluate in the \gtset{} setting.
As shown in \cref{tab:multilingual}, \methodname{} reaches 71.67 REF and 91.45 TIME, well ahead of MiMo-V2.6-Flash and Qwen3-Omni-Thinking, and completes every clip, whereas MiMo-V2.6-Flash falls into repetitive loops on 38 of 140 clips.
This suggests that \methodname{}'s diarization and timing generalize well across languages.

\begin{table}[!tp]
    \centering
    \footnotesize
    \setlength{\tabcolsep}{2pt}
    \begin{tabular}{@{}>{\raggedright\arraybackslash}p{\dimexpr\linewidth-11.0em-6\tabcolsep\relax}w{c}{5.5em}w{c}{2.5em}w{c}{3.0em}@{}}
        \toprule
        Method & \#Completed & REF$\uparrow$ & TIME$\uparrow$ \\
        \midrule
        MiMo-V2.6-Flash & 102/140 & 64.33 & 21.70 \\
        Qwen3-Omni-Thinking & 138/140 & 39.95 & 20.29 \\
        \rowcolor{gray!12}
        \methodname{} & 140/140 & \textbf{71.67} & \textbf{91.45} \\
        \bottomrule
    \end{tabular}
    \vspace{-0.5em}
    \caption{\textbf{Dialogue captioning in 14 languages} (\gtset{} setting). \#Completed excludes clips on which a model degenerates into repetitive loops, indicating that it handles the language poorly. {REF and TIME are computed on completed clips.}}
    \label{tab:multilingual}
    \vspace{-1em}
\end{table}

\noindent\textbf{Long videos.}
\label{sec:dvdbench}
DVDBench~\cite{tang2026dorca} contains 964 English clips of 84s on average and up to 5 minutes, with whole-second timestamps.
We split each video into 50--60s windows that overlap by 20s, caption them independently, and merge adjacent windows by linking speakers through their shared utterances, without any additional model.
The baselines instead process each clip in a single pass.
Compared with open-weight models, \methodname{} achieves the best Acc, WER, IoU, and REF, and trails only D-ORCA in TIME (\cref{tab:dvdbench}).
It matches Gemini~3.1~Pro in Acc (87.05 \vs 86.96) but trails both Gemini models in WER and REF, likely because each window is captioned without the rest of the video.

\begin{table}[!tbp]
    \centering
    \footnotesize
    \setlength{\tabcolsep}{2pt}
    \begin{tabular}{@{}>{\raggedright\arraybackslash}p{\dimexpr\linewidth-13.3em-10\tabcolsep\relax}w{c}{2.5em}w{c}{2.8em}w{c}{2.5em}w{c}{2.5em}w{c}{3.0em}@{}}
        \toprule
        Method & Acc$\uparrow$ & WER$\downarrow$ & IoU$\uparrow$ & REF$\uparrow$ & TIME$\uparrow$ \\
        \midrule
        \multicolumn{6}{@{}l}{\textit{Proprietary}} \\
        Gemini 3.1 Pro & 86.96 & \underline{15.96} & 49.56 & 69.82 & 56.87 \\
        Gemini 4 Argon & \underline{88.82} & 16.34 & \underline{62.86} & \underline{71.58} & \underline{65.36} \\
        \midrule
        \multicolumn{6}{@{}l}{\textit{Open-weight}} \\
        DiaDem & 80.46 & 23.32 & - & 62.50 & - \\
        D-ORCA & 80.00 & 19.07 & 56.80 & 62.45 & \textbf{63.28} \\
        Qwen3-Omni-Thinking & 76.86 & 20.62 & 42.14 & 58.33 & 49.65 \\
        \midrule
        \rowcolor{gray!12}
        \methodname{} & \textbf{87.05} & \textbf{16.68} & \textbf{61.18} & \textbf{69.34} & 62.03 \\
        \bottomrule
    \end{tabular}
    \vspace{-0.5em}
    \caption{\textbf{Long-video evaluation on DVDBench~\cite{tang2026dorca}.}
    Acc, WER, and IoU are the original DVDBench metrics, and TIME uses a 1s tolerance for its whole-second timestamps.
    \textbf{Bold} and \underline{underline} mark the best open-weight and proprietary results.}
    \label{tab:dvdbench}
    \vspace{-1.5em}
\end{table}

\begin{table}[!tbp]
    \centering
    \footnotesize
    \setlength{\tabcolsep}{2pt}
    \begin{tabular}{@{}>{\raggedright\arraybackslash}p{\dimexpr\linewidth-14.0em-8\tabcolsep\relax}w{c}{3.8em}w{c}{3.2em}w{c}{3.8em}w{c}{3.2em}@{}}
        \toprule
        & \multicolumn{2}{c}{Direct} & \multicolumn{2}{c@{}}{Caption first} \\
        \cmidrule(lr){2-3}\cmidrule(l){4-5}
        Method & Speaker$\uparrow$ & Audio$\uparrow$ & Speaker$\uparrow$ & Audio$\uparrow$ \\
        \midrule
        Transcript only & - & - & 37.2 & 33.7 \\
        Gemma 4 E4B & 31.7 & 38.2 & - & - \\
        Qwen3-Omni-Thinking & 54.0 & 56.5 & - & - \\
        video-SALMONN 2+ & 49.4 & 53.6 & - & - \\
        DiaDem & 52.0 & 52.7 & 59.1 & 56.9 \\
        D-ORCA & 55.5 & 55.1 & \textbf{62.7} & 58.8 \\
        \midrule
        \rowcolor{gray!12}
        \methodname{} & \textbf{58.2} & \textbf{62.0} & 59.0 & \textbf{67.1} \\
        \bottomrule
    \end{tabular}
    \vspace{-0.5em}
    \caption{\textbf{Audiovisual question answering (AV-QA) on AV-Speaker\-Bench~\cite{nguyen2026see}.}
    Accuracy on the speaker- and audio-centric questions.
    \textit{Direct}: each model answers from the video, audio, and question.
    \textit{Caption first}: each model generates a caption, from which Qwen3-Omni-Thinking answers the question.}
    \label{tab:avqa}
    \vspace{-1.5em}
\end{table}

\noindent\textbf{Audiovisual question answering (AV-QA).}
\label{sec:avqa}
We test whether the capabilities of \methodname{} transfer to AV-QA (\cref{tab:avqa}).
AV-SpeakerBench~\cite{nguyen2026see} contains 3,212 four-choice questions (\eg, speaker identification and temporal localization of speech) that cannot be answered from a single modality.
Although fine-tuned only for dialogue captioning, \methodname{} outperforms its base model, Qwen3-Omni-Thinking, by $+$4.2 Speaker and $+$5.5 Audio in the Direct setting.
In the Caption first setting, its captions yield the best Audio accuracy (67.1 \vs 58.8 for D-ORCA).
We also compare with \emph{Transcript only}, which gives Qwen3-Omni-Thinking only the speech transcript that \methodname{} takes as input.
\methodname{} outperforms it by $+$21.8 Speaker and $+$33.4 Audio, showing that its gains come from the model rather than the transcript.

\begin{figure}[!t]
    \centering
    \includegraphics[width=\columnwidth]{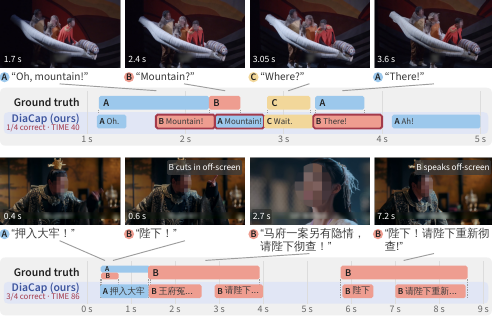}
    \vspace{-1.5em}
    \caption{\textbf{Failure cases.}
    \emph{Top:} small, distant faces cause segmentation and attribution errors.
    \emph{Bottom:} overlapping speech is missed.}
    \label{fig:qual-fail-clip991}
    \vspace{-1em}
\end{figure}

\vspace{-0.5em}
\section{Limitations}
\label{sec:limitations}
\vspace{-0.5em}

\emph{First}, \methodname{} needs stronger perception in challenging scenes.
For example, small, distant faces provide little lip-motion evidence, which causes \methodname{} to split a single utterance across two speakers in \cref{fig:qual-fail-clip991}.
\emph{Second}, \methodname{} would benefit from stronger reasoning, which could infer the speaker from the dialogue context when the visual evidence is unreliable, as in \cref{fig:qual-fail-clip991}.
\emph{Third}, transcription remains a bottleneck for diarization: REF increases by +9.35 for DiaCap given GT transcripts (\cref{tab:main_results}).
DiaCap-E2E with RL already matches the transcript-conditioned model without an external recognizer, and stronger transcription within the end-to-end model could narrow this gap further.

\vspace{-0.5em}
\section{Conclusion}
\label{sec:conclusion}
\vspace{-0.5em}

We presented \methodname{}, an audiovisual LLM for comprehensive dialogue captioning that learns utterance timing and audiovisual diarization jointly.
It is trained with dense supervision from 225K synthetically annotated clips and with fine-grained feedback from RL on 2.9K human-corrected clips.
Experimental results showed that \methodname{} achieves superior timestamp accuracy and strong audiovisual diarization, and \eename{} is comparably strong without an external recognizer.

\par

\clearpage
\appendix
\makeatletter
\setlength{\@dblfptop}{0pt}
\setlength{\@dblfpsep}{\dblfloatsep}
\setlength{\@dblfpbot}{0pt plus 1fil}
\makeatother

\section{SFT Corpus Statistics}
\label{app:data}

Since the filtered pool is unbalanced, we resample it to build the SFT corpus (\cref{sec:sft}).

For language, we keep every clip from the 16 underrepresented languages.
To prevent a few long videos from dominating, we keep at most one clip per video for English, Mandarin, and Hindi.
The underrepresented languages are exempt from this cap so that none of their clips are lost.

For the number of speakers, we keep every reachable clip with 3$+$ speakers.
When a long video has several candidate clips, we pick the one with 3$+$ speakers.
Two-speaker clips then fill the corpus until clips with 2$+$ speakers make up 64\% of it.
Single-speaker clips cannot drop much below 21\%, because the underrepresented languages, which we keep in full, are mostly single-speaker.
We also add about 16\% no-speech clips, which contain only ambient sound, music, sound effects, or silence.

\begin{figure}[!htbp]
    \centering
    \includegraphics[width=\columnwidth]{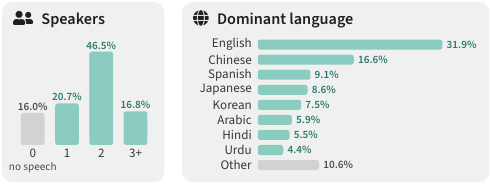}
    \caption{\textbf{SFT corpus statistics.}}
    \label{fig:data_statistics}
\end{figure}

The final corpus has 225K clips and
\cref{fig:data_statistics} shows the detailed distributions.

\section{Results on the Original Videos}
\label{app:original}

\cref{tab:joint_orig,tab:rl_data_orig,tab:reward_orig} repeat \cref{tab:joint,tab:rl_data,tab:reward} on DiaDemBench with the original videos, in which burned-in subtitles are not redacted, as in the \cmark{} columns of \cref{tab:main_results}.

\begin{table}[!htbp]
    \centering
    \footnotesize
    \setlength{\tabcolsep}{2.5pt}
    \begin{tabular}{llrrr}
        \toprule
        Method & Transcript & REF$\uparrow$ & ASR$\uparrow$ & TIME$\uparrow$ \\
        \midrule
        \eename{} & Self-predicted & {73.02} & {82.13} & {77.83} \\
        \midrule
        & Scribe v2 & 71.42 & 78.24 & 76.08 \\
        & \cellcolor{gray!15}Qwen3-Omni-Captioner$^*$ & \cellcolor{gray!15}71.87 & \cellcolor{gray!15}78.97 & \cellcolor{gray!15}76.56 \\
        & MAI-Transcribe-2 & 72.96 & 81.20 & \textbf{78.73} \\
        \multirow{-4}{*}{\methodname{}} & Gemini 4 Argon & \textbf{73.30} & \textbf{82.25} & 77.34 \\
        \bottomrule
    \end{tabular}
    \caption{\textbf{Transcript-conditioned \vs end-to-end prediction} on DiaDemBench with the original videos.
    Both \methodname{} and \eename{} are trained with SFT and RL, and other settings follow \cref{tab:joint}.
    $^*$Default recognizer. \textbf{Bold} marks the best results.}
    \label{tab:joint_orig}
\end{table}

\begin{table}[!htbp]
    \centering
    \footnotesize
    \setlength{\tabcolsep}{3pt}
    \begin{tabular}{llrrr}
        \toprule
        RL data & Labels & REF$\uparrow$ & ASR$\uparrow$ & TIME$\uparrow$ \\
        \midrule
        Random & Automatic & 70.88 & 78.66 & 75.85 \\
        Mined & Automatic & 71.34 & 78.92 & 76.31 \\
        \rowcolor{gray!15}
        Mined & Human-reviewed & \textbf{71.87} & \textbf{78.97} & \textbf{76.56} \\
        \bottomrule
    \end{tabular}
    \caption{\textbf{Effect of RL data} on DiaDemBench with the original videos.
    All variants start from the same SFT checkpoint and use our reward.
    \textbf{Bold} marks the best results.}
    \label{tab:rl_data_orig}
    \vspace{-1em}
\end{table}

\begin{table}[!htbp]
    \centering
    \footnotesize
    \setlength{\tabcolsep}{3pt}
    \begin{tabular}{lrrr}
        \toprule
        Method & REF$\uparrow$ & ASR$\uparrow$ & TIME$\uparrow$ \\
        \midrule
        SFT & 69.44 & 78.50 & 75.55 \\
        \;+RL w/ $0.8\,\mathrm{REF}{+}0.2R_{\text{time}}$ & 70.71 & 78.85 & 75.93 \\
        \rowcolor{gray!15}
        \;+RL w/ $0.4R_{\text{grp}}{+}0.4R_{\text{bind}}{+}0.2R_{\text{time}}$ & \textbf{71.87} & \textbf{78.97} & \textbf{76.56} \\
        \bottomrule
    \end{tabular}
    \caption{\textbf{Effect of the RL reward} on DiaDemBench with the original videos.
    Both rewards are trained on the same human-reviewed mined clips, and the shaded row is our reward.
    \textbf{Bold} marks the best results.}
    \label{tab:reward_orig}
    \vspace{-1em}
\end{table}

\begin{table}[!tbp]
    \centering
    \footnotesize
    \setlength{\tabcolsep}{2pt}
    \begin{tabular}{@{}>{\raggedright\arraybackslash}p{\dimexpr\linewidth-13.3em-10\tabcolsep\relax}w{c}{2.5em}w{c}{2.8em}w{c}{2.5em}w{c}{2.5em}w{c}{3.0em}@{}}
        \toprule
        Method & Acc$\uparrow$ & WER$\downarrow$ & IoU$\uparrow$ & REF$\uparrow$ & TIME$\uparrow$ \\
        \midrule
        \multicolumn{6}{@{}l}{\textit{Proprietary}} \\
        Gemini 3.1 Pro & 86.96 & \underline{15.96} & 49.56 & 69.82 & 56.87 \\
        Gemini 4 Argon & \underline{88.82} & 16.34 & \underline{62.86} & \underline{71.58} & \underline{65.36} \\
        \midrule
        \multicolumn{6}{@{}l}{\textit{Open-weight}} \\
        DiaDem & 80.46 & 23.32 & - & 62.50 & - \\
        D-ORCA & 80.00 & 19.07 & 56.80 & 62.45 & 63.28 \\
        Qwen3-Omni-Thinking & 76.86 & 20.62 & 42.14 & 58.33 & 49.65 \\
        \midrule
        \rowcolor{gray!12}
        \methodname{} & \textbf{87.05} & 16.68 & \textbf{61.18} & 69.34 & 62.03 \\
        \rowcolor{gray!12}
        \methodname{}-E2E & 86.72 & \textbf{15.89} & 60.60 & \textbf{69.46} & \textbf{63.81} \\
        \bottomrule
    \end{tabular}
    \vspace{-0.5em}
    \caption{\textbf{Long-video evaluation on DVDBench.}
    Acc, WER, and IoU are the original DVDBench metrics, and TIME uses a 1s tolerance for its whole-second timestamps.
    \textbf{Bold} and \underline{underline} mark the best open-weight and proprietary results.}
    \label{tab:dvdbench-w-e2e}
    \vspace{-1em}
\end{table}

\begin{table*}[!htbp]
    \centering
    \footnotesize
    \setlength{\tabcolsep}{3.5pt}
    \begin{tabular}{@{}l*{11}{c}@{}}
        \toprule
         & \multicolumn{7}{c}{Audio-centric} & \multicolumn{4}{c}{Speaker-centric} \\
        \cmidrule(lr){2-8}\cmidrule(l){9-12}
        Method & Count$\uparrow$ & Duration$\uparrow$ & Intensity$\uparrow$ & Pitch$\uparrow$ & Rate$\uparrow$ & ASR$\uparrow$ & All$\uparrow$ & Count$\uparrow$ & Detect$\uparrow$ & Recog$\uparrow$ & All$\uparrow$ \\
         & (288) & (236) & (206) & (206) & (209) & (201) & (1,346) & (402) & (427) & (422) & (1,251) \\
        \midrule
        \multicolumn{12}{@{}l}{\textit{Direct: the model answers from the video, audio, and question}} \\
        Gemma 4 E4B & 34.0 & 39.0 & 34.5 & 42.2 & 37.3 & 43.8 & 38.2 & 29.1 & 34.2 & 31.8 & 31.7 \\
        Qwen3-Omni-Thinking & \textbf{40.3} & 58.5 & 67.5 & 57.8 & 52.2 & 69.7 & 56.5 & \textbf{45.8} & 56.7 & 59.2 & 54.0 \\
        video-SALMONN 2+ & 29.5 & 59.7 & 54.4 & 57.8 & 56.5 & 73.1 & 53.6 & 39.1 & 55.7 & 52.8 & 49.4 \\
        DiaDem & 29.5 & 58.9 & 59.7 & 56.8 & 49.3 & 70.6 & 52.7 & 43.5 & 56.9 & 55.2 & 52.0 \\
        D-ORCA & 38.5 & 62.3 & 54.4 & 49.0 & 57.9 & 74.6 & 55.1 & 42.5 & \textbf{63.9} & 59.2 & 55.5 \\
        \rowcolor{gray!12}
        \methodname{} & 37.2 & \textbf{68.6} & 70.9 & 68.4 & \textbf{61.7} & 74.1 & \textbf{62.0} & 44.3 & 63.5 & \textbf{66.1} & \textbf{58.2} \\
        \rowcolor{gray!12}
        \methodname{}-E2E & 34.4 & 65.7 & \textbf{74.3} & \textbf{69.9} & 61.2 & \textbf{75.6} & 61.7 & 41.0 & 63.7 & 64.0 & 56.5 \\
        \midrule
        \multicolumn{12}{@{}l}{\textit{Caption first: the model writes a caption, from which Qwen3-Omni-Thinking answers}} \\
        Transcript only & 36.5 & 25.4 & 23.3 & 35.0 & 31.1 & 51.7 & 33.7 & 31.3 & 49.6 & 30.3 & 37.2 \\
        DiaDem & 52.8 & 54.2 & 59.7 & 51.0 & 62.2 & 63.7 & 56.9 & 56.5 & 64.9 & 55.7 & 59.1 \\
        D-ORCA & 53.8 & 63.6 & 57.3 & 54.9 & 52.2 & \textbf{72.6} & 58.8 & \textbf{57.7} & \textbf{68.1} & \textbf{61.8} & \textbf{62.7} \\
        \rowcolor{gray!12}
        \methodname{} & 59.7 & 67.8 & 62.6 & \textbf{66.5} & \textbf{77.0} & 71.6 & 67.1 & 51.5 & 63.7 & 61.4 & 59.0 \\
        \rowcolor{gray!12}
        \methodname{}-E2E & \textbf{63.9} & \textbf{70.8} & \textbf{63.1} & \textbf{66.5} & 69.9 & 70.1 & \textbf{67.2} & 51.5 & \textbf{68.1} & 61.1 & 60.4 \\
        \bottomrule
    \end{tabular}
    \caption{\textbf{AV-QA on AV-SpeakerBench~\cite{nguyen2026see} by task.}
    Accuracy (\%) on the audio- and speaker-centric tasks, with question counts in parentheses.
    \textit{All} pools the questions of each category.
    \textit{Direct}: each model answers from the video, audio, and question.
    \textit{Caption first}: each model writes a question-blind caption, from which Qwen3-Omni-Thinking answers.
    \textbf{Bold} marks the best result in each block.}
    \label{tab:avqa_tasks}
    \vspace{-1.2em}
\end{table*}

\noindent\textbf{Transcript conditioning.}
\cref{tab:joint_orig} repeats the comparison in \cref{tab:joint} on the original videos.
Two findings from the redacted videos still hold.
First, \eename{} achieves strong results.
With the default Qwen3-Omni-Captioner, it now outperforms \methodname{} in all three metrics, by $+$1.15 REF, $+$3.16 ASR, and $+$1.27 TIME.
Second, \methodname{} can use a different recognizer without retraining, and a stronger recognizer yields a better caption.
With Gemini~4~Argon, \methodname{} reaches the best REF (73.30) and ASR (82.25), both higher than those of \eename{}.
With MAI-Transcribe-2, it reaches the best TIME (78.73).

These results also support evaluating on redacted videos, as in the main paper.
When burned-in subtitles are visible, a model that sees the video can read the words off the screen instead of recognizing them from the audio.
This changes which recognizer comes out ahead.
Gemini~4~Argon is the only recognizer in \cref{tab:joint_orig} that sees the video, and with it \methodname{} now achieves a higher ASR than with the audio-only MAI-Transcribe-2 (82.25 \vs 81.20), the reverse of the redacted results.
\eename{} also sees the video and gains in the same way.
Its ASR lead over \methodname{} with the audio-only Qwen3-Omni-Captioner grows from 0.56 on the redacted videos to 3.16 here.
Part of the score on the original videos therefore comes from reading subtitles, which favors models that see the video but reveals little about how well a model links speech to speakers.
Redacting the subtitles removes this shortcut, so every model must transcribe from the audio alone.

\noindent\textbf{RL data.}
The trends in \cref{tab:rl_data} hold on the original videos (\cref{tab:rl_data_orig}).
With the same automatic labels, mined clips outperform random ones by $+$0.46 REF and $+$0.46 TIME.
Human review of the mined labels adds a further $+$0.53 REF and $+$0.25 TIME, for a total gain of $+$0.99 REF and $+$0.71 TIME over random clips with automatic labels.

\noindent\textbf{RL reward design.}
The trends in \cref{tab:reward} also hold on the original videos (\cref{tab:reward_orig}).
Our reward outperforms the REF-based baseline by $+$1.16 REF and $+$0.63 TIME.
It improves REF over SFT by $+$2.43 instead of $+$1.27 for the baseline, even though the baseline optimizes REF directly.

\section{Additional Generalization Results}
\label{app:generalization}

\cref{sec:generalization} evaluates \methodname{} beyond DiaDemBench on long videos and audiovisual question answering.
This section evaluates \eename{} on the same two benchmarks.

\subsection{Long Videos}
\label{app:dvdbench}

\cref{tab:dvdbench-w-e2e} adds \eename{} to \cref{tab:dvdbench} on DVDBench.
Both variants share the overlapping windows and cross-window speaker linking of \cref{sec:generalization}, and differ only in that \eename{} transcribes each window itself instead of taking the transcript from a speech recognizer.
\eename{} matches \methodname{} in REF (69.46 \vs 69.34) and improves TIME by $+$1.78 (63.81 \vs 62.03), the best open-weight result, ahead of D-ORCA (63.28).
Its WER of 15.89 is the lowest of all models, including Gemini~3.1~Pro (15.96) and Gemini~4~Argon (16.34), while \methodname{} keeps a small lead in Acc (87.05 \vs 86.72) and IoU (61.18 \vs 60.60).
As on DiaDemBench (\cref{tab:main_results}), \eename{} is as strong as \methodname{}, here on videos of up to five minutes.

\subsection{Audiovisual Question Answering by Task}
\label{app:avqa}

\begin{figure*}[t]
  \centering
  \includegraphics[width=\textwidth]{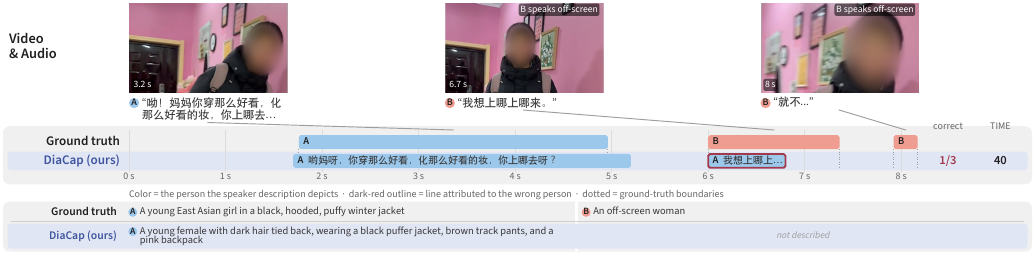}
  \caption{\textbf{Failure case: off-screen speaker merged with the on-screen one.}
  In a handheld phone video, a girl (A) asks her mother where she is going, and the mother (B) answers from off-screen.
  \methodname{} describes a single speaker, gives the mother's reply (``I'll go wherever I want'') to the girl, and drops the mother's cut-off last words.
  Since only the girl is ever on screen, the model does not infer that a second person is speaking.}
  \label{fig:qual-fail-clip928}
\end{figure*}

\cref{tab:avqa_tasks} reports the per-task result of AV-SpeakerBench~\cite{nguyen2026see} in the two settings of \cref{tab:avqa}.
In the Direct setting, \methodname{} and \eename{} improve over their base model, Qwen3-Omni-Thinking, on both the audio-centric ($+$5.5 and $+$5.2) and the speaker-centric averages ($+$4.2 and $+$2.5). They also outperform D-ORCA on both audio-centric ($+$6.9 and $+$6.6) and speaker-centric ($+$2.7 and $+$1.0) tasks.
In the Caption first setting, they give the best audio-centric accuracy (67.1 and 67.2 \vs 58.8 for D-ORCA).

\noindent\textbf{\eename{}.}
\eename{} performs on par with \methodname{} in both settings.
Answering directly, it is within 0.3 and 1.7 points of \methodname{} on the audio- and speaker-centric tasks (61.7 \vs 62.0 and 56.5 \vs 58.2).
As a captioner, it matches \methodname{} on the audio-centric tasks (67.2 \vs 67.1) and is $+$1.4 higher on the speaker-centric tasks (60.4 \vs 59.0).

\noindent\textbf{Audio-centric tasks.}
In the Direct setting, \methodname{} and \eename{} improve over the base model on five of the six audio-centric tasks (duration, intensity, pitch, rate, and speech recognition), by up to $+$10.6 and $+$12.1 on pitch.
These tasks ask how and when each person speaks, which our captions describe with the timestamps, words, and voice description of every utterance, so this skill carries over to question answering although both models are fine-tuned only to write captions.
\eename{}, which also learns to transcribe, is the best model on speech recognition (75.6).
The exception is speech counting, where both models fall behind the base model (37.2 and 34.4 \vs 40.3), as its questions ask how many times a phrase is said, often by one person within an interval marked by a visual event, which a model answering directly must tally without writing anything down.
When the models first write out every utterance with its speaker (Caption first), their accuracy rises to 59.7 and 63.9, and \eename{} is the best in that block.

\noindent\textbf{Speaker-centric tasks.}
In the Direct setting, the largest gain is on speaker recognition, where \methodname{} improves over the base model by $+$6.9 (66.1 \vs 59.2) and is the best model.
These questions ask who says a given line or who speaks right before or after an event, which requires linking each utterance to the visible person who says it, the core of audiovisual diarization that our SFT and RL train.
On speaker counting, \methodname{} is $-$1.5 behind the base model (44.3 \vs 45.8).
Each answer gives the number of both speakers and people on screen, and \methodname{} errs mostly on the latter by undercounting partly visible people, a visual skill outside our training.
On speaker detection, which asks whether a person speaks before or after an event, \methodname{} improves over the base model by $+$6.8 but is $-$0.4 behind D-ORCA (63.5 \vs 63.9).
Like its base model, it tends to answer yes and correctly answers 78 of the 185 questions whose answer is no (\vs 96 for D-ORCA).

\section{Additional Failure Case}
\label{app:failure}

\cref{fig:qual-fail-clip928} shows another failure case, in which an off-screen speaker is merged with the only person on screen.
As in \cref{fig:qual-fail-clip991}, each \methodname{} bar shows the predicted speaker and text, and a ground-truth utterance with no bar below it was not predicted.
Handling overlapping speech and correcting recognition errors are directions for future work.

\section{Additional Qualitative Results}
\label{app:qualitative}

\begin{figure*}[t]
  \centering
  \includegraphics[width=\textwidth]{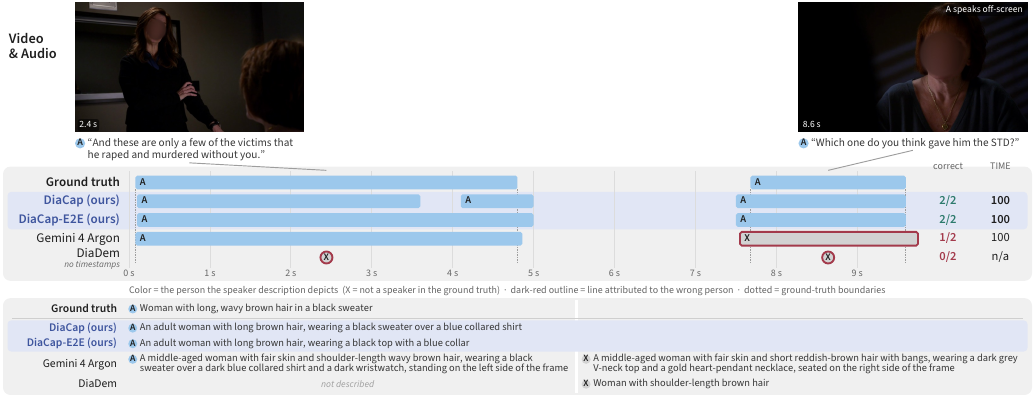}
  \caption{\textbf{One speaker who goes off-screen.}
  In this typical shot/reverse-shot conversation, a single woman speaks both lines.
  During the second line, the camera cuts to the listener and the speaker is off-screen.
  \methodname{} and \eename{} attribute both lines to the same speaker.
  Gemini~4~Argon gives the second line to the visible listener (short reddish-brown hair).
  DiaDem's description (``woman with shoulder-length brown hair'') is too generic to identify the speaker, so neither of its lines is credited.}
  \label{fig:qual_clip101}
\end{figure*}

\begin{figure*}[t]
  \centering
  \includegraphics[width=\textwidth]{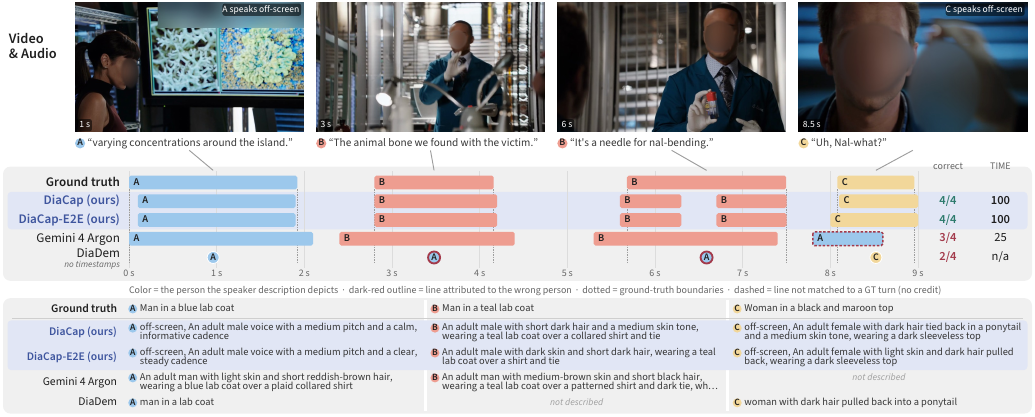}
  \caption{\textbf{Off-screen speakers in a multi-shot scene.}
  The first and last turns are spoken while the camera shows someone else.
  \methodname{} and \eename{} attribute all four turns correctly and describe these speakers as off-screen voices.
  Gemini~4~Argon gives the woman's ``Uh, Nal-what?'' to the man in the blue lab coat and truncates it.
  DiaDem describes both men only as ``man in a lab coat'', so the second man's two turns go to the first man.}
  \label{fig:qual_clip199}
\end{figure*}

\begin{figure*}[t]
  \centering
  \includegraphics[width=\textwidth]{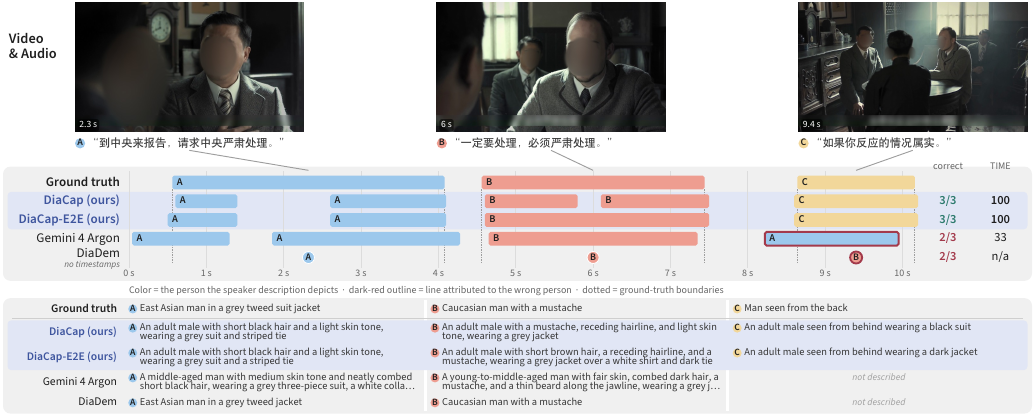}
  \caption{\textbf{A speaker seen only from behind.}
  The last line is spoken by the man with his back to the camera.
  \methodname{} and \eename{} attribute it to him.
  Gemini~4~Argon gives it to the man in the gray suit and DiaDem to the man with the mustache, the two people whose faces are visible.}
  \label{fig:qual_clip341}
\end{figure*}

\begin{figure*}[t]
  \centering
  \includegraphics[width=\textwidth]{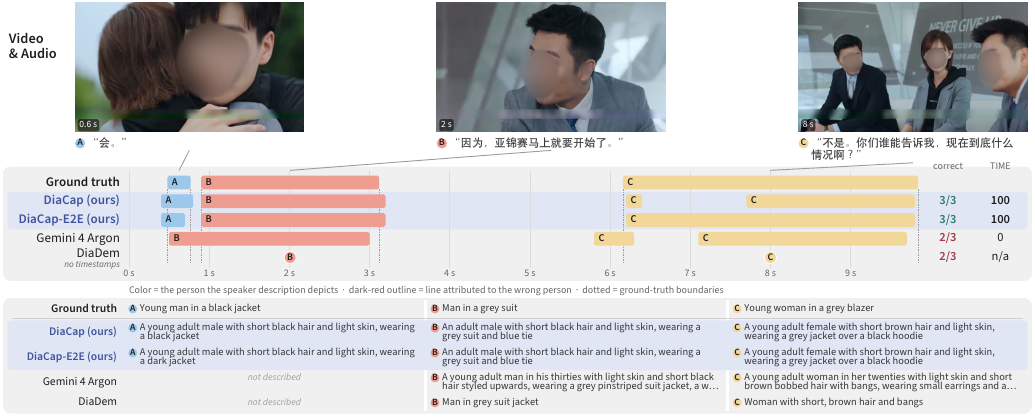}
  \caption{\textbf{A one-word turn right before another speaker (Mandarin).}
  The young man answers with a single syllable lasting 0.3s (first frame), and the man in the gray suit continues immediately (second frame).
  \methodname{} and \eename{} keep the short turn as a separate line with the correct speaker.
  Gemini~4~Argon and DiaDem merge it into the next speaker's line, so turn A is lost.
  Gemini's timestamps also miss every boundary (TIME 0), and DiaDem predicts none.
  This example shows that accurate timestamps and diarization are tightly coupled, as a model without precise turn boundaries tends to absorb a short turn into a longer neighboring one.}
  \label{fig:qual_clip125}
\end{figure*}

\begin{figure*}[t]
  \centering
  \includegraphics[width=\textwidth]{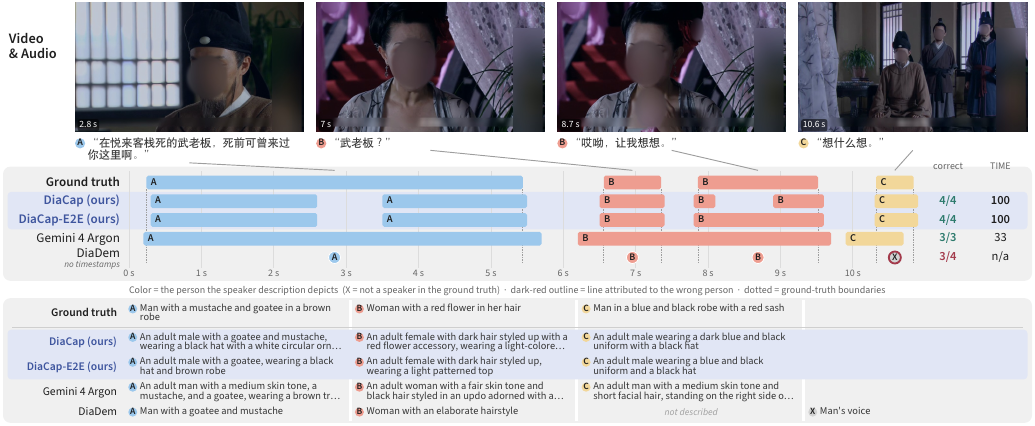}
  \caption{\textbf{A short interjection from a third speaker.}
  \methodname{} and \eename{} attribute all four turns correctly and place every boundary within the collar.
  Gemini~4~Argon merges the woman's two turns into one line, so when scoring Gemini, we merge consecutive ground-truth utterances from the same speaker, which leaves three ground-truth turns.
  Gemini's timestamps miss most boundaries (TIME 33).
  DiaDem describes the speaker of the final interjection only as ``man's voice'', which depicts none of the people in the scene.}
  \label{fig:qual_clip276}
\end{figure*}

\begin{figure*}[t]
  \centering
  \includegraphics[width=\textwidth]{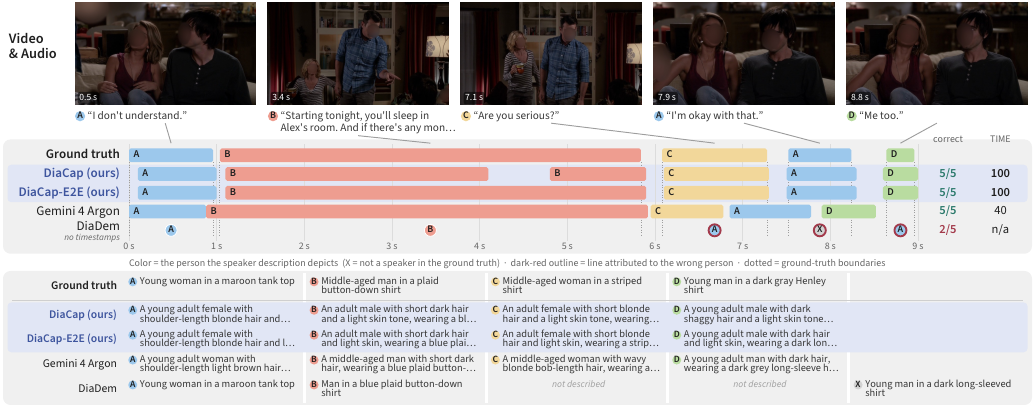}
  \caption{\textbf{Four speakers in one room.}
  \methodname{}, \eename{}, and Gemini~4~Argon attribute all five turns correctly.
  DiaDem gets three of the five turns wrong.
  It gives ``Are you serious?'' (C) and ``Me too.'' (D) to the young woman, and gives her own ``I'm okay with that.'' to a young man.
  Gemini's timestamps drift by about 0.7s on the last three turns (TIME 40), whereas both \methodname{} variants place every boundary within the collar (TIME 100).}
  \label{fig:qual_clip48}
\end{figure*}

Figures~\ref{fig:qual_clip101}--\ref{fig:qual_clip48} show more examples in the layout of \cref{fig:case_study}.
Each figure shows one keyframe per ground-truth utterance, labeled with the utterance spoken at that moment.
The timeline compares the ground truth with \methodname{}, \eename{}, Gemini~4~Argon~\cite{google2026gemini4}, and DiaDem~\cite{chen2026diadem}.
Each bar is a predicted utterance placed at its predicted timestamps and colored by the ground-truth speaker that the model's description depicts (X marks a person who never speaks), and a dark-red outline marks a wrong speaker.
DiaDem predicts no timestamps, so its utterances are drawn as dots under the ground-truth utterances they are matched to.
For each model, \emph{correct} counts the ground-truth utterances attributed to the right speaker, and TIME is the clip-level TIME score.
\methodname{} runs in the \predset{} setting, and the baselines are end-to-end.
The panel below each timeline lists the visual part of each model's speaker descriptions.

\noindent\textbf{Speakers who are not visible.}
In conversation scenes, the camera often shows the listener rather than the speaker (\cref{fig:qual_clip101,fig:qual_clip199}), or the speaker faces away from the camera (\cref{fig:qual_clip341}).
Baselines tend to assign such lines to whoever is visible, whereas \methodname{} and \eename{} link the voice to the right person across shots.

\noindent\textbf{Short turns in fast exchanges.}
A one-word answer or a brief interjection between longer turns is easily absorbed into a neighboring turn (\cref{fig:qual_clip125,fig:qual_clip276}).
\methodname{} and \eename{} keep such turns separate with accurate boundaries, whereas baselines with imprecise or no timestamps merge them, which loses the turn and its speaker.

\noindent\textbf{Several similar-looking speakers.}
With three or more speakers in one scene, a description must tell each person apart (\cref{fig:qual_clip48,fig:qual_clip199}).
DiaDem's generic descriptions, such as ``man in a lab coat'', fit more than one person and lead to swapped or merged speakers, whereas the descriptions of \methodname{} remain specific enough to identify each speaker.

\FloatBarrier
\clearpage
{
    \small
    \bibliographystyle{ieeenat_fullname}
    \bibliography{main}
}

\end{document}